\documentclass[letterpaper, 10pt, conference]{IEEEtran}
 
\usepackage{siunitx}
\usepackage{times}
\usepackage{graphicx}
\usepackage{amsmath}
\usepackage{amssymb}
\usepackage{algorithm}
\usepackage{algorithmic}
\usepackage{booktabs}
\usepackage{multirow}
\usepackage{cite}
\usepackage{url}
\usepackage{xcolor}

\IEEEoverridecommandlockouts
 
\title{\LARGE \bf Learning-Based Hierarchical Scene Graph Matching for\\Robot Localization Leveraging Prior Maps}

\author{
  Nimrod Millenium Ndulue$^{1*}$,
  Jose Andres Millan-Romera$^{1*}$,
  Matteo Giorgi$^{2}$,
  Holger Voos$^{1}$,
  Jose Luis Sanchez-Lopez$^{1}$%
\thanks{$^{*}$ These authors contributed equally to this work.}
\thanks{$^{1}$Authors are with the Automation and Robotics Research Group, Interdisciplinary Centre for Security, Reliability and Trust (SnT), University of Luxembourg. Holger Voos is also associated with the Faculty of Science, Technology and Medicine, University of Luxembourg, Luxembourg.
\tt{\small{\{nimrodmillenium.ndulue, jose.millan, holger.voos, joseluis.sanchezlopez\}}@uni.lu}}
\thanks{$^{2}$Author is with the University of Pisa.
\tt{\small{m.giorgi13@studenti.unipi.it}}}
\thanks{This work was partially funded by the Fonds National de la Recherche of Luxembourg (FNR) under the projects 17097684/RoboSAUR and C22/IS/17387634/DEUS.}
\thanks{For the purpose of Open Access, and in fulfillment of the obligations arising from the grant agreement, the authors have applied a Creative Commons Attribution 4.0 International (CC BY 4.0) license to any Author Accepted Manuscript version arising from this submission.}
}
\begin{document}
 
\maketitle
\thispagestyle{empty}
\pagestyle{empty}
 
\begin{abstract}
Accurate localization is a fundamental requirement for autonomous robots
operating in indoor environments. Scene graphs encode the spatial
structure of an environment as a hierarchy of semantic entities and
their relationships, and can be constructed both online from robot
sensor data and offline from architectural priors such as Building
Information Models (BIM). Matching these two complementary
representations enables drift correction in SLAM by grounding robot
observations against a known structural prior. However, establishing
reliable node-to-node correspondences between them remains an open
challenge: existing combinatorial methods are prohibitively expensive
at scale, and prior learned approaches address only flat graph
matching, ignoring the multi-level semantic structure present in both
representations. Here we present a learned, end-to-end differentiable
pipeline that augments both graphs with semantically motivated edge
types encoding intra- and inter-level relationships, explicitly
exploiting this hierarchy to enable simultaneous matching from
high-level room concepts down to low-level wall surfaces. Trained
exclusively on floor plans, the proposed method outperforms the
combinatorial baseline in F1 on real LiDAR environments while running
an order of magnitude faster, demonstrating viable zero-shot
generalization for BIM-assisted robot localization.
\end{abstract}
 
\section{Introduction}
\label{sec:intro}

Scene graph-based SLAM has gained significant traction as a powerful 
paradigm for robots operating in indoor environments, enriching metric 
reconstruction with a hierarchical representation of semantic spatial 
entities --- such as rooms and wall surfaces --- and their topological 
relationships~\cite{hughes2022hydra, sgraph}. These entities are 
heterogeneous in nature and exist at different levels of abstraction, 
posing a challenge for matching methods that assume homogeneous node 
representations. Yet sensor noise and drift remain fundamental limitations: 
without an external reference, accumulated errors corrupt both the robot's 
localization capability and the consistency of its scene graph over time. 
In indoor environments, Building Information Models (BIM) provide exactly 
such a reference, encoding the geometric and semantic structure of buildings 
as structured prior knowledge available before any robot exploration 
begins~\cite{isgraph}. Leveraging BIM to constrain scene graph-based SLAM 
is therefore a natural and promising direction, and has been shown to 
substantially improve localization accuracy by grounding robot observations 
against stable architectural features~\cite{bimsemanticlocalization}.

The central problem is therefore to establish node-to-node correspondences
between the scene graph built by the robot and an equivalent graph derived
from the BIM: finding such correspondences not only corrects accumulated
drift but also allows the robot to align its scene graph with the
architectural prior, jointly improving localization and the consistency
of its semantic map. This problem is further complicated by the fact that 
the robot typically operates under partial observations, having explored 
only a portion of the environment, so its scene graph is in general 
smaller and incomplete with respect to the BIM-derived one. Despite its 
importance, this problem remains open. Prior work has progressed from 
dense metric registration against BIM geometry~\cite{bimslam} to matching 
semantic scene features~\cite{semanticlocbim}, with the state of the art 
casting this as a hierarchical graph matching problem solved via 
combinatorial search~\cite{isgraph}. However, combinatorial methods do 
not scale to real-time deployment, and while learned approaches have 
demonstrated superior scalability and generalization over classical 
combinatorial solvers for flat, homogeneous graph 
matching~\cite{ngm}, none address the matching problem in the presence 
of heterogeneous node types across multiple semantic levels.

We address this with a learned, end-to-end differentiable pipeline that
exploits the two-level hierarchy shared by both graphs.
Our contributions are:
\begin{itemize}
\item A graph augmentation strategy that enriches both graphs with two 
additional edge types encoding intra- and inter-level spatial 
relationships, enabling information flow across both hierarchy levels 
within a single unified graph.
\item A shared Multi-Layer Perceptron (MLP) encoder that maps the 
heterogeneous node features of rooms and wall surfaces into a common 
embedding space, producing type-aware and directly comparable node 
representations.
\item A learned end-to-end differentiable pipeline for hierarchical scene
graph matching across high-level (room) and low-level (wall) entities
under partial observations, transferring zero-shot to real LiDAR 
environments.
\end{itemize}
\section{Related Work}
\label{sec:related}

\subsection{Architectural Prior-Based Robot Localization}
Early approaches localize robots by registering raw sensor data against
metric maps derived from architectural plans~\cite{semanticlocbim, bimslam, liobim},
but dense geometric alignment scales poorly and is sensitive to
plan-reality deviations. None of these methods exploit the topological
relationships between architectural entities, limiting their ability to
reason about structural organization.

\subsection{Graph Matching}
Graph matching seeks to establish node correspondences between two
graphs. Classical formulations cast this as a quadratic assignment
problem~\cite{ngm}, which is NP-hard and thus intractable for large
graphs. Learned approaches overcome this by combining graph neural network (GNN)-based
embeddings with differentiable Sinkhorn
assignment~\cite{pcagm, cie, common}. With
SuperGlue~\cite{superglue} and SG-PGM~\cite{sgpgm} bringing this
paradigm to robotics for keypoint matching and 3D scene graph
matching respectively. However, these methods operate on flat graphs of homogeneous nodes, not designed to exploit hierarchical structure between node types of fundamentally different geometric nature.
iS-Graphs~\cite{isgraph} addresses this by performing hierarchical metric-semantic graph matching between a robot-built scene graph and a BIM-derived architectural prior, yet its combinatorial search grows exponentially with the number of rooms and factorially with the number of walls, making it prohibitively expensive at scale.

\section{System Overview}
\label{sec:system_overview}

\begin{figure*}[t]
    \centering
    \includegraphics[width=0.9\textwidth]{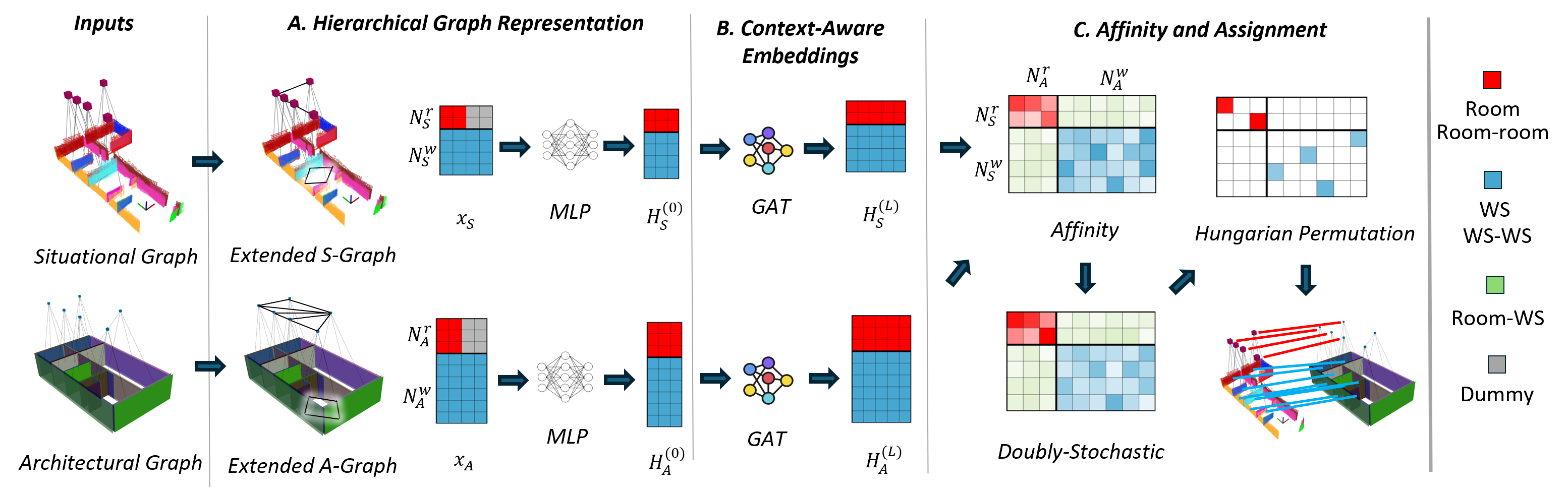}
    \caption{\textbf{Overview of the proposed pipeline.} A shared MLP improves the initial node features, after which a shared GATv2 encoder
    produces structure-aware embeddings for both the A-graph (derived
    from BIM) and the S-graph (built online from LiDAR SLAM). A
    dot-product affinity matrix is computed, normalized via Sinkhorn
    with dummy-column padding to handle partial observations, and
    decoded into a hard one-to-one correspondence by the Hungarian
    algorithm.}
    \label{fig:pipeline}
\end{figure*}

The proposed system takes as input an Architectural Graph
(A-graph)~\cite{isgraph} derived from BIM and a Situational Graph
(S-graph)~\cite{sgraph} built online from robot observations, and
produces a node-to-node correspondence between them. Since the robot
may have only partially explored the environment, the S-graph is typically smaller than the A-graph, and the matching is formulated as a
rectangular bipartite assignment. As shown in Fig.~\ref{fig:pipeline},
the pipeline encodes both graphs, scores cross-graph node similarity
through an affinity matrix, and produces a soft assignment matrix that
is decoded into a hard one-to-one correspondence at inference. The
following subsections detail each component.
 
\subsection{Hierarchical Graph Representations}
\label{sec:graphs}

Both the A-graph~\cite{isgraph} and the S-graph~\cite{sgraph} share a
common two-level hierarchical structure: \emph{rooms}, representing
enclosed spaces, and \emph{wall surfaces} (WS), representing the
planar surfaces that delimit them.

\paragraph{Node features.}
Each node is described by a feature vector $x_i \in \mathbb{R}^7$, see
Tab.~\ref{tab:node_features}.

\begin{table}[h]
\centering
\caption{Node feature vector composition for room and WS nodes.}
\label{tab:node_features}
\begin{tabular}{clll}
\hline
\textbf{Dims} & \textbf{Feature} & \textbf{Room} & \textbf{WS} \\
\hline
1--2 & Node type (one-hot) & $[1,0]$ & $[0,1]$ \\
3--4 & 2D centroid $(x, y)$ & $(x_r, y_r)$ & $(x_w, y_w)$ \\
5--6 & 2D surface normal $(n_x, n_y)$ & $(0, 0)$ & $(n_x, n_y)$ \\
7    & Segment length & $-1$ & $\ell$ \\
\hline
\end{tabular}
\end{table}

Dimensions (1--2) encode the node type as a one-hot vector
(\texttt{room}$=[1,0]$, \texttt{ws}$=[0,1]$); (3--4)
encode the 2D centroid coordinates; (5--6) encode the
2D outward-facing surface normal; (7) encodes the segment
length, set to $-1$ for room nodes, which carry no meaningful
length attribute. Node features are standardized using the mean and standard deviation computed on the training set and applied consistently across all splits.

\paragraph{Feature homogenization.}
As shown in Tab.~\ref{tab:node_features}, the proposed graph is
heterogeneous, since room and WS nodes are characterized by different
meaningful feature dimensions. To make these features comparable across
node types, each raw node feature vector $x_i$ is
transformed by a two-layer MLP into an initial node embedding of higher
dimension, allowing a more expressive representation to be learned
before message passing. Each layer applies an affine transformation
followed by a ReLU activation and dropout:
\begin{equation}
    \phi^{(l)}(z) = \mathrm{Dropout}\!\left(\mathrm{ReLU}\!\left(\mathbf{W}^{(l)} z + \mathbf{b}^{(l)}\right)\right)
    \label{eq:mlp_layer}
\end{equation}
where $\mathbf{W}^{(l)}$ and $\mathbf{b}^{(l)}$ are the learnable
parameters of layer $l$. The initial node embedding is then computed as
\begin{equation}
    h_i^{(0)} = \phi^{(2)}(\phi^{(1)}(x_i)).
    \label{eq:node_embedding_init}
\end{equation}

\paragraph{Edge structure.}
Classic formulations of both the A-graph and the S-graph connect each
room node only to the walls that belong to
it~\cite{isgraph,sgraph}. To enrich the structural information available
to the encoder, we augment both graphs with two additional edge types,
illustrated in Fig.~\ref{fig:pipeline}.A in black.
The full edge set therefore comprises three directed relation types.
\emph{Room-to-WS edges} capture spatial containment, connecting each
room to its constituent wall segments, the only edge type present in
the original formulations. \emph{Room-to-room edges} encode adjacency
between spaces that share a physical wall. \emph{WS-to-WS edges} connect
wall segments belonging to the same room in an angularly-sorted ring
around the room center. Together, these augmented edge types encode
spatial context across both abstraction levels within a single unified
graph. This richer connectivity allows the GNN to propagate
context both within and across hierarchy levels during message passing,
which would not be possible with the original single-edge-type graphs.
 
\subsection{Context-Aware Node Embeddings via GATv2}
\label{sec:encoder}
 
Raw node features are enriched through GATv2~\cite{gatv2}, whose
pairwise attention mechanism computes scores over the joint representation
of both edge endpoints, allowing the model to learn distinct notions of
relevance for neighbors of fundamentally different geometric nature ---
such as a WS node attending to its parent room versus its angular siblings.
The attention score for each directed edge $(u \rightarrow v)$ is:
 
\begin{equation}
e_{uv} = \mathbf{a}^\top \,\text{LeakyReLU}
         \!\left(\mathbf{W}_a \cdot [\mathbf{h}_u \| \mathbf{h}_v]\right)
\label{eq:attention_score}
\end{equation}
 
where $\mathbf{W}_a$ and $\mathbf{a}$ are learnable parameters. The
resulting attention weights $\alpha_{uv}$ update each node representation
as:
 
\begin{equation}
\mathbf{h}_v^{(l+1)} = \sum_{u \in \mathcal{N}(v)}
                        \alpha_{uv} \cdot \mathbf{W}_h\mathbf{h}_u^{(l)}
\label{eq:node_update}
\end{equation}
 
The encoder consists of $L{=}2$ GATv2 layers with 4 attention heads,
hidden dimension 64, and output embedding dimension 32. ReLU and node
dropout ($p{=}0.15$) are applied after the first layer only; attention
dropout ($p{=}0.12$) is applied at both. The encoder is shared across
both graphs, projecting rooms and wall segments into a common embedding
space where cross-graph similarity is meaningful.
 
\subsection{Cross-Graph Affinity and Differentiable Assignment}
\label{sec:affinity}

Building on the affinity-Sinkhorn pipeline introduced in
PCA-GM~\cite{pcagm} and adapted to scene graph matching in
SG-PGM~\cite{sgpgm}, node-to-node correspondences are established
by computing a pairwise affinity matrix and solving the resulting
combinatorial assignment through a continuous relaxation over
soft assignment matrices.

Let $\mathbf{H}_1^{(L)} \in \mathbb{R}^{N_1 \times d}$ and $\mathbf{H}_2 ^{(L)}\in
\mathbb{R}^{N_2 \times d}$ denote the node embeddings of the A-graph
($N_1$ nodes) and S-graph ($N_2$ nodes), with $d$ the embedding dimension.
The affinity matrix is computed as:

\begin{equation}
    \mathbf{A} = \mathbf{H}_1 \mathbf{H}_2^\top \in \mathbb{R}^{N_1 \times N_2}
\end{equation}

where $A_{ij}$ measures the similarity between node $i$ in the A-graph
and node $j$ in the S-graph. A plain dot product is used in place of the
learnable bilinear form of PCA-GM~\cite{pcagm}, which was found to cause
gradient explosions in preliminary experiments~\cite{giorgi2024}. Instance
Normalization is then applied to $\mathbf{A}$, standardizing each sample
to zero mean and unit variance to ensure well-conditioned input to the
Sinkhorn layer across graphs of variable size.

Because the S-graph is in general smaller than the A-graph
($N_2 \leq N_1$), $\mathbf{A}$ is rectangular and Sinkhorn cannot be
applied directly. To recover a square matrix, $N_1 - N_2$ dummy columns
are appended to $\mathbf{A}$, yielding $\hat{\mathbf{A}} \in
\mathbb{R}^{N_1 \times N_1}$. The dummy columns absorb the probability mass of A-graph nodes 
that have no observed counterpart in the S-graph, providing a 
principled mechanism to handle graph matching when the affinity 
matrix is rectangular, without requiring explicit outlier rejection. The Sinkhorn algorithm~\cite{sinkhorn} is
then applied to $\hat{\mathbf{A}}$ to approximate the one-to-one
correspondence constraint: by iteratively normalizing rows and columns
to sum to one, each node is forced to distribute its total probability
mass across candidates in the other graph, concentrating it on the most
compatible matches, producing a doubly-stochastic matrix
$\mathbf{S} \in [0,1]^{N_1 \times N_1}$. At inference, the $N_1 \times N_2$ submatrix
of $\mathbf{S}$ corresponding to real S-graph nodes is passed to the
Hungarian algorithm~\cite{hungarian}, extracting a globally optimal
hard one-to-one correspondence in $\mathcal{O}(n^3)$.

\section{Experimental Evaluation}
\label{sec:experiments}

\subsection{Setup}
\label{sec:setup}

\begin{figure}[t]
    \centering
    \includegraphics[width=0.9\linewidth]{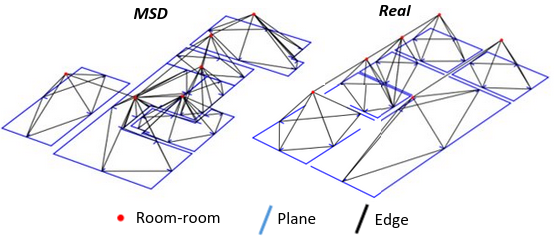}
    \caption{Example graph  used for evaluation: a floor plan from the MSD
    synthetic dataset (left) and the real indoor environment RE with its
    corresponding A-graph constructed from architectural blueprints (right).}
    \label{fig:realBIM}
\end{figure}

\textbf{Datasets.}
We evaluate the proposed method on both synthetic and real data.
For \textsc{Synthetic} evaluation, the MSD dataset~\cite{msd}
provides \textit{real} architectural floor plans represented as A-graphs. S-graphs
are simulated by applying perturbations to these A-graphs, mimicking
the partial and noisy observations a robot builds online. We consider
the \textit{ws-room noise} condition, in which wall surface and room
nodes are randomly dropped and geometric noise is applied.
For \textsc{Real} evaluation, an S-graph was acquired through a Velodyne VLP-16 LiDAR sensor in a real indoor
environment (RE), see Fig.~\ref{fig:realBIM}. The corresponding
A-graph was constructed from architectural blueprints. No retraining
or fine-tuning was performed, making this a zero-shot generalisation
test.

\textbf{Baselines.}
We compare against iS-Graphs~\cite{isgraph}, the only prior work
performing hierarchical semantic graph matching between S-graphs and
A-graphs. To the best of our knowledge, no learned method addresses
this problem, making iS-Graphs the most relevant baseline.

\textbf{Evaluation metrics.}
Matching is evaluated as binary classification over all $N_2 \times N_1$
candidate node pairs, where $N_2$ is the number of S-graph nodes and
$N_1$ is the number of A-graph nodes, as defined in
Section~\ref{sec:affinity}. We report \textbf{Precision},
\textbf{Recall}, \textbf{F1}, \textbf{Accuracy}, and \textbf{Matching
time}. F1 is the primary metric of interest, as true negative dominance
inflates accuracy and makes it a misleading measure of matching quality.

\subsection{Training}
\label{sec:training}

The model is trained on the MSD dataset (Section~\ref{sec:setup}), split
into training (\SI{70}{\percent}), validation (\SI{15}{\percent}), and
test (\SI{15}{\percent}) subsets using a stratified split that preserves
the distribution of graph sizes across all sets.

The loss function is the Permutation Loss~\cite{pcagm}, a Binary
Cross-Entropy applied element-wise between the predicted soft assignment
matrix $\mathbf{S}$ and the ground-truth assignment matrix
$\mathbf{P}_{\mathrm{gt}}$. BCE penalizes both false positives ---
node pairs incorrectly predicted as matches --- and false negatives ---
true correspondences missed by the model --- independently across all
entries, better suiting the sparse bijective structure of the assignment
problem than row-wise Cross-Entropy. Optimization is performed with AdamW (learning rate $1{\times}10^{-3}$, weight decay $5{\times}10^{-5}$, batch size 16). Early stopping on validation loss is applied to prevent overfitting.
Hyperparameters were tuned using Optuna~\cite{optuna} with the
Tree-structured Parzen Estimator (TPE) sampler.
 
 
\subsection{Results and Discussion}
\label{sec:results}

\textbf{MSD dataset.}
Table~\ref{tab:results_synthetic} reports matching performance on the
MSD test split, comparing the proposed method against
iS-Graphs~\cite{isgraph}. Due to the factorial and exponential
complexity of iS-Graphs, a timeout threshold was applied: samples
exceeding 60 seconds are discarded, and the percentage completed
within the timeout is reported. While iS-Graphs achieves higher
precision and F1 on completed samples, only 86\% of samples finish
within the timeout, confirming that its combinatorial cost does not
scale to real-time deployment. On the other side, the proposed method
processes all samples deterministically, with competitive matching
quality. In terms of computation time, the proposed method achieves an
$82\times$ speedup over iS-Graphs (0.093\,s vs.\ 7.66\,s per sample),
consistent with the 14\% of samples that exceed the timeout threshold
and are discarded from iS-Graphs' evaluation.

\begin{table}[h]
\centering
\caption{Matching performance on the MSD synthetic test set.
$^\dagger$Percentage of samples completed within the timeout.}
\label{tab:results_synthetic}
\begin{tabular}{lccccc}
\toprule
\textbf{Method} & \textbf{Prec.\%} & \textbf{Rec.\%} &
\textbf{F1\%} & \textbf{Time(s)} & \textbf{Completed$^\dagger$} \\
\midrule
iS-Graphs~\cite{isgraph} & 99 & 92 & 95 & 7.66 & 86\% \\
Ours                     & 85 & 85 & 85 & 0.093 & 100\% \\
\bottomrule
\end{tabular}
\end{table}

\textbf{Real dataset.}
Table~\ref{tab:results_real} reports performance on the real
environment (RE). Despite being trained exclusively on synthetic
data, the proposed method outperforms iS-Graphs in F1. iS-Graphs
achieves higher precision but at the cost of recall: its conservative
combinatorial search only commits to matches satisfying strict
geometric constraints, leaving many true correspondences unmatched
under real sensor noise. The GNN-based matcher, trained to penalise
both false positives and false negatives, learns a more flexible
assignment that better tolerates domain shift, yielding higher recall
and overall F1. Notably, iS-Graphs completes within the timeout here
due to the relatively low number of nodes in RE. Nevertheless, the
proposed method remains $9\times$ faster (0.021\,s vs.\ 0.191\,s per
sample); on larger environments, where node count increases, its
combinatorial cost would render it impractical.

\begin{table}[h]
\centering
\caption{Matching performance and inference time on real LiDAR
environment RE. Best results in \textbf{bold}.}
\label{tab:results_real}
\begin{tabular}{lccccc}
\toprule
\textbf{Method} & \textbf{Prec.\%} & \textbf{Rec.\%} &
\textbf{F1\%} & \textbf{Time [s]} & \textbf{Completed} \\
\midrule
iS-Graphs~\cite{isgraph} & \textbf{88} & 53 & 67 & 0.191 & 100\% \\
Ours & 80 & \textbf{88} & \textbf{84} & \textbf{0.021} & 100\% \\
\bottomrule
\end{tabular}
\end{table}


\section{Conclusion}
\label{sec:conclusion}

We presented a learned, end-to-end differentiable pipeline for
hierarchical scene graph matching between BIM-derived Architectural
Graphs and robot-built Situational Graphs, achieving competitive
matching quality while running orders of magnitude faster than the combinatorial baseline and transferring zero-shot to real LiDAR environments. A current limitation is that the model does not account for structural symmetry, which may lead to ambiguous correspondences in symmetric floor plans. As future work, we plan to first improve precision and recall of the learned matcher, and subsequently address symmetry detection in environments with structurally similar rooms and repeated wall configurations.
 
\bibliographystyle{IEEEtran}
\bibliography{references}
 
\end{document}